\documentclass[lettersize,journal]{IEEEtran}
\usepackage{amsmath,amsfonts}
\usepackage{algorithmic}
\usepackage{array}
\usepackage[caption=false,font=normalsize,labelfont=sf,textfont=sf]{subfig}
\usepackage{textcomp}
\usepackage{stfloats}
\usepackage{url}
\usepackage{verbatim}
\usepackage{graphicx}
\usepackage{cite}
\usepackage{booktabs}
\usepackage{multirow}
\def\BibTeX{{\rm B\kern-.05em{\sc i\kern-.025em b}\kern-.08em
    T\kern-.1667em\lower.7ex\hbox{E}\kern-.125emX}}
\usepackage{balance}

\begin{document}
\title{FER-former: Multi-modal Transformer for Facial Expression Recognition}
\author{Yande Li, Mingjie Wang, Minglun Gong, Yonggang Lu, Li Liu
\thanks{%This work was supported by grands from China Scholarship Council (Grant No. 202006180033) and was done when Yande Li studied at the University of Guelph. (\emph{Corresponding author: Minglun Gong; Yonggang Lu; Li Liu}.)\par
Yande Li is with the School of Information Science and Engineering, Lanzhou University, Lanzhou, 730000, China and also with the School of Computer Science, University of Guelph, Guelph, N1G 2W1, Canada (e-mail: liyd19@lzu.edu.cn). \par
Mingjie Wang is with the School of Science, Zhejiang Sci-Tech University, Hangzhou, 310018, China (e-mail: mingjiew@zstu.edu.cn). \par
Minglun Gong is with the School of Computer Science, University of Guelph, Guelph, N1G 2W1, Canada (e-mail: minglun@uoguelph.ca). \par
Yonggang Lu is with the School of Information Science and Engineering, Lanzhou University, Lanzhou, 730000, China (e-mail: ylu@lzu.edu.cn). \par
Li Liu is with the School of Big Data and Software Engineering, Chongqing University, Chongqing, 401331, China (e-mail: dcsliuli@cqu.edu.cn).}}

\markboth{Journal of Arxiv}%
{How to Use the IEEEtran \LaTeX \ Templates}

\maketitle

\begin{abstract}
The ever-increasing demands for intuitive interactions in Virtual Reality has triggered a boom in the realm of Facial Expression Recognition (FER).
To address the limitations in existing approaches (e.g., narrow receptive fields and homogenous supervisory signals) and further cement the capacity of FER tools, a novel multifarious supervision-steering Transformer for FER in the wild is proposed in this paper.
Referred as FER-former, our approach features multi-granularity embedding integration, hybrid self-attention scheme, and heterogeneous domain-steering supervision.
In specific, to dig deep into the merits of the combination of features provided by prevailing CNNs and Transformers, a hybrid stem is designed to cascade two types of learning paradigms simultaneously.
Wherein, a FER-specific transformer mechanism is devised to characterize conventional hard one-hot label-focusing and CLIP-based text-oriented tokens in parallel for final classification.
To ease the issue of annotation ambiguity, a heterogeneous domains-steering supervision module is proposed to make image features also have text-space semantic correlations by supervising the similarity between image features and text features.
On top of the collaboration of multifarious token heads, diverse global receptive fields with multi-modal semantic cues are captured, thereby delivering superb learning capability. Extensive experiments on popular benchmarks demonstrate the superiority of the proposed FER-former over the existing state-of-the-arts.
\end{abstract}

\begin{IEEEkeywords}
Annotation ambiguity, CLIP, facial expression recognition, multi-modal, vision Transformer.
\end{IEEEkeywords}

%%%%%%%%% BODY TEXT

\section{Introduction}
Recently, Facial Expression Recognition (FER) has carved a promising research path in computer vision toward fine cognition for human facial expressions as they are one of most natural and effective carriers to express emotions.
Impressive performance \cite{li2022dual,yang2020novel} has been attained on strictly lab-controlled datasets (e.g., MMI \cite{valstar2010induced}, CK+ \cite{lucey2010extended}, and Oulu-CASIA \cite{zhao2011facial}), where all samples were produced and annotated in a heuristically-defined manner with low annotation error. However, the transferability of these models to scenes in the wild have large room for improvements.
Compared to FER in controlled environments, FER in the wild is more challenging due to the issues of limited samples, annotation ambiguity, and unconstrained variations (occlusion, pose, illumination, etc.).
Yet, the FER's capability of adapting for natural scenarios is of paramount importance for the prospective generalization/deployment of FER models in practical applications.

Convolutional Neural Networks (CNNs) have achieved roaring success for various vision problems such as semantic segmentation \cite{zhou2022contextual}, object detection \cite{wang2021end}, and image classification \cite{sun2021supervised}.
Since Kahou et al. \cite{kahou2013combining} and Tang et al. \cite{tang2013deep} unprecedentedly adopt CNNs into FER and win the EmotiW 2013 and FER 2013 emotion recognition competitions, CNNs-based approaches become the dominating techniques and a pool of pioneering methods have been developed.
For example, to ameliorate the occlusion problem, several current studies \cite{wang2020region,LiZSC19,zhao2021learning} attempt to weaken the negative impacts of occlusion regions in raw images by delving into more sophisticated CNNs architectures.
Albeit the promising achievements and insights, these CNNs-based methods only focus on the merits of CNNs (e.g., pyramidal representations and well-established pretrained models.), but overlook the limitations of narrow receptive fields. Lacking global contextual semantics easily impels the model to produce error-prone classification results and hinders the generalization of FER algorithms. To enlarge the receptive fields of features, several existing works have strived to deepen the networks or improve convolution path \cite{han2019enhanced,gu2017enlarging}.
Nevertheless, these approaches still follow the learning paradigm of CNNs and representations with global semantics are neglected.

In recent couple of years, the learning paradigm of transformer, with self-attention mechanisms, has demonstrated its advantages among learning-based models in Natural Language Processing (NLP) \cite{vaswani2017attention, devlin2018bert}. It has also been widely applied in computer vision to model global relationships among pixels over 2D feature maps. In particular, Vision Transformer (ViT) \cite{dosovitskiy2020image} is a pioneering work and becomes favoured baseline stem for various vision tasks, e.g., semantic segmentation \cite{gu2022multi}, image classification \cite{sun2022spectral}, object detection \cite{han2022few}, and image processing \cite{tu2022maxim}.
Although global receptive fields can be modelled via self-attention condensers, the insufficiency of training data \cite{dosovitskiy2020image} impedes the training of models from scratch, because the pure transformer block often requires a larger amount of training data. Besides, patch-based tokens learned by transformer loss the pyramidal structure of features, which damages the extraction of local spatial information. To circumvent these challenges, TransFER \cite{xue2021transfer} is proposed to combine both pre-trained CNNs-based IR-50 and Transformer-based encoder, resulting intriguing FER performance. However, the collaboration procedure is coarse in exiting hybrid approaches and they somewhat ignore the fine-grained extraction/integration of multi-scale features, thereby reducing the robustness and scale invariance of representations. Hence, there is still large room for further improvements of FER algorithms to absorb the merits of hybrid CNNs and Transformer paradigms in one pass.

Another common nuisance in the wild FER is that only manually-annotated one-hot ground truths (hard) are provided in most of existing prevailing datasets, which easily incurs annotation ambiguity due to the subjectiveness of annotators, ambiguous facial expressions, or low-quality facial images. For example, Chen et al. \cite{chen2021understanding} demonstrate the existence of annotation biases between \emph{genders} in many expression datasets, whereas the work \cite{mao2021label} suggests that inter-correlations among emotions are inherently disparate, consequently degrading the performance of FER.

Lots of efforts have been made to address the issue of annotation ambiguity and can be roughly categorized as two mainstream directions: Learning of Soft Label Distribution and Learning from Hard Samples via delicately-designed mechanisms (e.g., attention mechanism or self-supervised strategy). Specifically, \emph{i)} Instead of hard one-hot supervisory signals, a set of probability distributions with auxiliary priors are assigned to facial images with the goal of driving the deep learning-based models to extract representations with fuzzy boundaries \cite{zhao2021robust,chen2020label,she2021dive,shao2022self}; \emph{ii)} Hard samples are selected as outliers and then imposed on extra special processes to suppress their negative impacts on the holistic models' learning \cite{shao2022self,wang2020suppressing,Li2022towards,wang2022ease}. Even though a series of attempts have been conducted to ease annotation ambiguity, the required auxiliary supervisory signals are still derived from raw one-hot labels and the ambiguity is prone to be inherited. Hence, homogenous supervision with intrinsic ambiguity is considered by existing endeavours and the merits of heterogeneous supervisory signals from multi-modal annotations are grossly underestimated or ignored. It is a remarkable fact that the recent resounding success of CLIP \cite{radford2021learning} allows the thriving development of various multi-modal learning (involving image and text) in many fields \cite{conde2021clip, wang2022clip}, and therefore provides a strong inspiration for our research.

Motivated by the above observations, in this paper, we propose a novel \emph{Multifarious Supervision-steering Transformer} for FER in the wild, dubbed as FER-former, which features multi-granularity embedding integration, hybrid self-attention scheme and  heterogeneous supervisory signals (one-hot and self-defined textual labels).
To promote effective collaboration between features from prevailing CNNs (pyramidal property) and Transformer (global receptive fields), a hybrid structure is designed for possessing both learning paradigms in one pass.
To mitigate the issues of uncontrolled occlusion and pose variations, a downsampling scheme imposed on the crude features from pre-trained CNN stem is devised to disentangle diverse spatial cues, while increasing the scale invariance of tokens.
%Besides,
%a multi-scale strategy is also proposed to increase the scale invariance of tokens.
%In specific, raw features are split into sequences of patches with varied resolutions and projected as multi-scale tokens.
To ease the annotation ambiguity and boost the diversity of supervisory signals, we introduce textual annotations using natural language for each emotion, and then treat them as the auxiliary heterogeneous supervision to work together with the conventional hard one-hot labels.
On top of these two types of signals, a FER-specific transformer block containing a hybrid self-attention scheme (\emph{e.g. one-hot and text-focusing heads}) is delicately devised in our FER-Former to produce the ambiguity-robust tokens.
In summery, the main contributions are fourfold in this paper.

\begin{itemize}
	\item A novel heterogeneous supervision-steering Transformer for FER in the wild is proposed, namely FER-former, which features multi-granularity embedding integration, hybrid self-attention scheme and heterogeneous domain-steering supervision.
	\item A FER-specific transformer block that involves a hybrid self-attention scheme is delicately devised to characterize conventional one-hot and text-oriented tokens for heterogeneous classifications, where text semantic information is passed to class token and other patch tokens by steering token in each encoder block.
    \item To ease the issue of annotation ambiguity, a heterogeneous domains-steering supervision module is proposed to make image features also have text-space semantic correlations by supervising the similarity between image features and text features.
	\item Extensive experiments on popular benchmarks demonstrate the superiority of the proposed FER-former over the existing state-of-the-art approaches.
\end{itemize}

%------------------------------------------------------------------------
\section{Related work}
In this section, we review related studies which can be roughly summarized into three categories: FER in the wild, hybrid leaning paradigms, and vision-language pretraining.

\subsection{FER in the wild}
\paragraph{FER with unconstrained variations}
To handle FER under weaker input constraints, researchers have mainly focused on two challenges: occlusion and non-frontal head pose. To suppress their negative impacts on FER models' performance, remarkable efforts have been made.
Li et al. \cite{LiZSC19} split intermediate feature maps into 24 sub-patches, and then utilize self-attention mechanism to recalibrate the patch-based information for occlusion-aware FER. %on the basis of facial landmarks
Methods in \cite{pan2019occluded, xia2020occluded} propose to leverage non-occluded facial images as auxiliary information to further fine-tune the occluded network, thereby producing occlusion-robust representations for FER.
To increase the pose invariance of FER models, Zhang et al. \cite{zhang2020geometry} try to disentangle expression and head pose from a given image by using geometry information.
In addition, several works address the challenges of occlusion and pose variations simultaneously. For example,
Wang et al. \cite{wang2020region} crop five fixed regions from a raw image and then feed them into a region attention network (RAN) for modelling inner-relationships among these facial regions. Zhao et al. \cite{zhao2021learning} propose a global multi-scale and local attention network (MA-Net) with different kernel sizes and attention-based recalibration weights.
Even though various attempts have been exerted to widen the receptive fields, they all based on the CNNs learning paradigm and do not make full use of the global semantic features of facial images.

\paragraph{FER with annotation ambiguity}
Annotation ambiguity is inevitable in wild expression datasets due to the subjectiveness of annotators, ambiguous facial expressions, and low-quality facial images \cite{wang2020suppressing}.
Label distribution learning is an intuitive and favoured scheme to reduce the ambiguity.
Zhao et al. \cite{zhao2021robust} treat the output of an auxiliary ResNet-50 as probability distribution to guide the learning of the other backbone network.
Shao et al. \cite{shao2022self} further adopt an auxiliary network as a label distribution generator to generate label distributions for guiding the backbone network training and selecting easy samples.
Chen et al. \cite{chen2020label} put forward an approach to learn label distribution by calculating the confidence probability of its nearest neighbours. Mao et al. \cite{mao2021label} try to incorporate the correlations among expressions into the label distribution in the semantic space.
Besides, DMUE \cite{she2021dive} makes use of auxiliary multi-branching framework to mine the latent distribution while estimating the uncertainty of the annotation ambiguity.
%In addition, many other excellent approaches\cite{wang2020suppressing, Li2022towards, wang2022ease} try to alleviate the annotation ambiguity by absorbing extra informative clues from hard samples.

Many excellent approaches try to alleviate the annotation ambiguity by absorbing extra informative clues from hard samples. Wherein, Wang et al. \cite{wang2020suppressing} propose the combination of attention mechanism, ranking regularization and relabelling low-ranked samples to evade the influence of uncertain samples on the final recognition results. Ada-CM \cite{Li2022towards} is proposed to learn adaptive confidence margin for each expression and exploit the low-confidence samples. In \cite{wang2022ease}, Wang et al. divide all training samples into three groups: clean, noisy, and conflict, and then focus on learning with ambiguous samples to enhance the robustness of models.
Despite the impressive results have been achieved by existing methods through reconstructing label distribution or emphasizing on hard samples, they still derive the additional supervisory signals from raw one-hot labels and the ambiguity is inclined to be inherited.
In this paper, to mitigate the issues of annotation ambiguity, a FER-specific transformer block is well devised to characterize heterogeneous hard one-hot label-focusing and text-oriented classification tokens.

\subsection{Hybrid leaning paradigms}
Recently, ViT \cite{dosovitskiy2020image} has become a dominating component for various computer vision models thanks to its capability of modelling long-range dependencies among a set of sub-patches. Even though transformer has higher ability of capturing global receptive fields, the intrinsic merits of CNNs (e.g., pyramidal feature maps) are beneficial for enhancing model capacity. Hence, a series of hybrid architectures are designed to inherit the advantages of both CNNs and Transformer-based learning paradigms. For example,
CMT \cite{guo2022cmt} and CeiT \cite{yuan2021incorporating} combine depth-wise convolution with a feed-forward network in encoder blocks to compensate for the limitations of pure transformers.
The proposed convolutional token embedding and convolutional projection in CVT \cite{wu2021cvt} allow the model to inherit the merits of CNNs while enlarging the receptive fields via transformer. Moreover, LeViT \cite{graham2021levit} devise a fabric-like neural network for fast inference image classification by incorporating some components that were proven useful for convolutional architectures (replace the uniform structure of a Transformer by a pyramid with pooling). BoTNet \cite{srinivas2021bottleneck} achieves promising performance on image classification, object detection, and instance segmentation by replacing the spatial convolutions with global self-attention mechanism in the final three bottleneck blocks of a ResNet.

When handling FER tasks, the priori knowledge of pre-trained models are necessary for good performance due to limited datasets. Inspired by hybrid structures in other vision tasks, TransFER \cite{xue2021transfer} combines a pre-trained IR50 and a pre-trained Transformer encoder to accomplish a state-of-the-art FER performance. However, the collaboration procedure is coarse and somewhat ignores the extraction/integration of multi-scale features, thereby reducing the robustness and scale invariance of representations. Hence, there is still large room for fully exploiting the merits of hybrid CNNs and Transformer paradigms in one pass.

\subsection{Vision-language pretraining}
Vision-language pretraining (VLP) aims to learn strong representations from large-scale image-text pairs, which has achieved impressive performance on a variety of downstream vision-language tasks \cite{li2021align,radford2021learning}.
Most current approaches in VLP can be divided into two categories: fusion encoder(s) and dual encoders \cite{singh2022flava}.
The first category aims to model the shared self-attention between image features and text features with complex multimodal encoders, such as ALBEF \cite{li2021align}, FLAVA \cite{singh2022flava}, SIMVLM \cite{wang2021simvlm}, UniT \cite{hu2021unit}, ViLT \cite{kim2021vilt} and VinVL \cite{zhang2021vinvl}.
These methods are capable of handing complex tasks such as object detection, natural language understanding, and multimodal reasoning, but they mostly require pre-trained object detectors and high-resolution input images, and their small corpus limits their ability to generalize to specific fine-grained classification tasks.
In contrast, the methods in the second category such as ALIGN \cite{jia2021scaling} and CLIP \cite{radford2021learning} focus on learning separate unimodal encoders for image and text, which are simple and good at retrieval tasks.
In particular, CLIP \cite{radford2021learning} was trained from scratch on a large dataset of 400 million (image, text) pairs collected from the internet, which supports a wide range of corpus requirements for fine-grained classification tasks.
It shows the superiority of text supervision on multiple vision tasks \cite{conde2021clip, wang2022clip}, thus provides a strong inspiration for our research.

%-------------------------------------------------------------------------
\section{Methods}

\begin{figure*}
	\centering
	\includegraphics[width=1\textwidth]{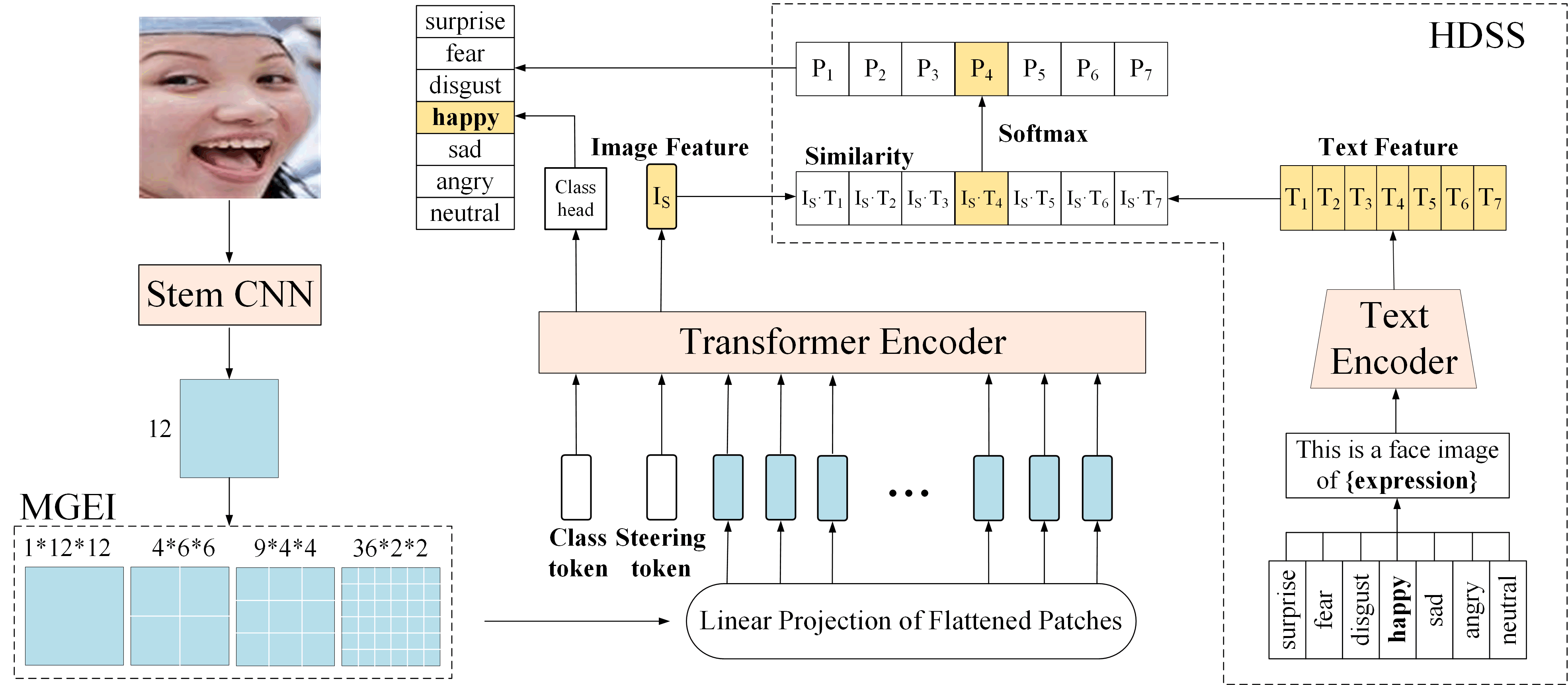}
	\caption{Pipeline of the proposed model. It consists of multi-granularity embedding integration (MGEI), hybrid self-attention, and heterogeneous domains-steering supervision (HDSS). The diverse global receptive fields with multi-modal semantic cues are captured by integrating text semantics into multi-head self-attention (MHSA). For simplicity, the convolutional position encodings are omitted.}
	\label{fig1}
\end{figure*}

\subsection{Overview}
An overview of the model is depicted in Fig \ref{fig1}. The proposed FER-former consists of multi-granularity embedding integration (MGEI), hybrid self-attention, and heterogeneous domains-steering supervision (HDSS) modules.
Same as the previous study in \cite{xue2021transfer}, the IR-50 \cite{deng2019arcface} pre-trained on Ms-Celeb-1M \cite{guo2016ms}) is adopted as a feature extractor due to its good generalization.
The pre-trained text encoder in CLIP \cite{radford2021learning} is fine-tuned to perceive the rich relationship among different text labels.

A scheme of downsampling the crude features from pre-trained CNN stem is devised to disentangle diverse spatial cues, yielding one global patch and 49 local patches of different scales. Then, all the patches are flattened and projected as multi-scale tokens.
%Unlike standard transformer, the position embedding is not necessary in our FER-former.
A novel FER-specific transformer encoder block containing a hybrid self-attention scheme is devised to characterize heterogeneous supervision signals.
More importantly, the extracted text features from the text encoder is delicately used to realize heterogeneous domains-steering
supervision by calculating image-text similarity.
More details are illustrated as follows.

\subsection{Multi-granularity embedding integration (MGEI)}
As described earlier, ViT does not generalize well when trained on insufficient amounts of data.
Therefore, to make full use of the merits of CNNs and ViT, a pre-trained stem network is utilized as a feature extractor, obtaining deep features $ X \in \mathbb{R} ^{C\times H\times W}$  where $C$, $H$ and $W$ refer to the number of channels, height, and width.
To facilitate splitting the extracted features into sequences of patches with various resolutions, the extracted features $X$ are pooled into $ X^{'} \in \mathbb{R} ^{C\times 12\times 12}$.
The pooled features are then downsampled to obtain diverse spatial cues that enable the model to overcome the issues of occlusion and pose variances.
Let $ X_{p_{i}} \in \mathbb{R} ^{N_{i}\times C \times P_{i}^{2}}, i\in (1,4)$ denote the local patches of different resolutions, where $N_{i}$ represents the patch numbers of each resolution, $(P_{i},P_{i})$ is the resolution of each feature patch.
To satisfy standard Transformer Encoder that receives a $1D$ sequence of token embeddings as input, all the patches are flattened and projected to uniform embeddings, which is denoted as $ X_{p} \in \mathbb{R} ^{N \times D}$, where $N = \sum_{i=1}^{4} N_{i}$ is the number of all patches and $D$ is the length of each feature embedding.

\subsection{Hybrid self-attention}
\begin{figure}[h]
	\centering
	\includegraphics[width=0.5\textwidth]{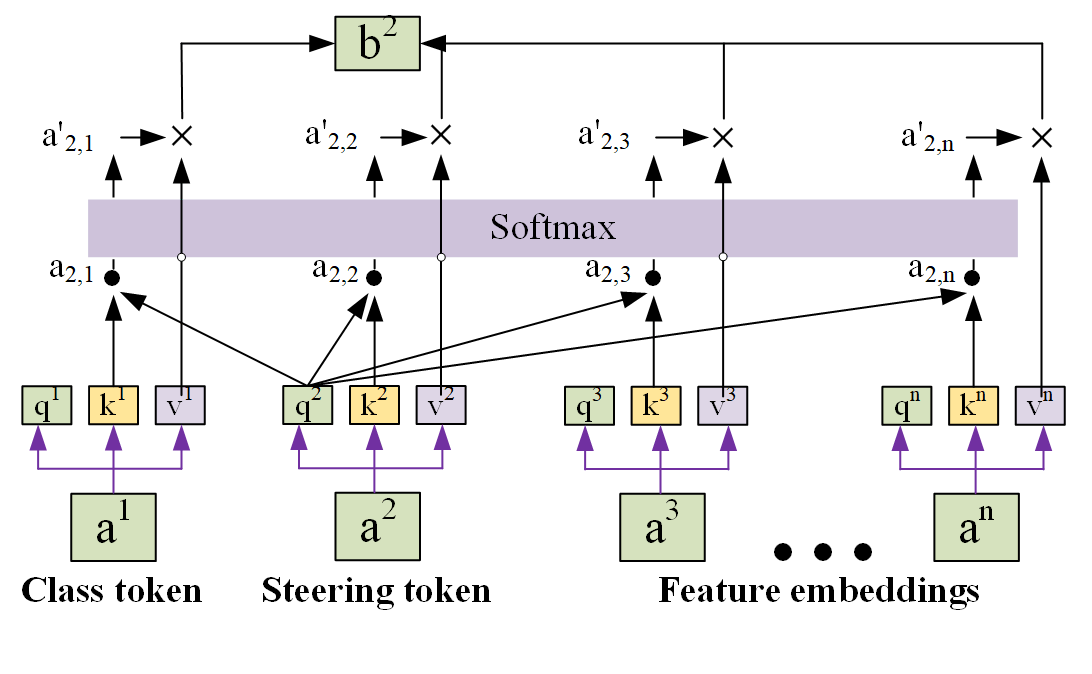}
	\caption{Flowchart of hybrid self-attention scheme. The last layer's class token is directly utilized for ground truth supervision, while the newly defined steering token is utilized for the integration of image and text information by calculating their similarity.}
	\label{fig2}
\end{figure}

The standard transformer encoder consists of a stack of $M$ encoder blocks, and each block contains alternating layers of multiheaded self-attention and multi-layer perception.
As a key to ViT's success, the self-attention makes it possible to embed information globally across the whole image.
As shown in Fig \ref{fig2}, we propose a novel hybrid self-attention that can characterize heterogeneous hard one-hot label-focusing and text-oriented classification tokens, where the text semantic information is passed to class token and other patch tokens by steering token in each encoder block.

Given a sequence of $1D$ feature embeddings $X_{p} \in \mathbb{R} ^{N \times D}$ and a class token $X_{c} \in \mathbb{R} ^{1 \times D}$, the routine input of an encoder block can be denoted as $Z^{'} = [X_{c}; X_{p}^{1}; X_{p}^{2}; \ldots; X_{p}^{N}]$.
To integrate image features with text features, a hybrid self-attention mechanism is designed.
Specifically, a steering token $X_{s} \in \mathbb{R} ^{1 \times D}$ is defined, yielding a new input for our FER-specific transformer encoder, which we denote as $Z = [X_{c}; X_{s}; X_{p}^{1}; X_{p}^{2}; \ldots; X_{p}^{N}]$.
For simplicity, we rename it as $Z = [a^{1}; a^{2}; \ldots; a^{n}]$, where $n = N + 2$. The steering token $a^{2}$ is connected with all other embeddings by calculating self-attention in each encoder block.
Firstly, the input $Z$ is linearly transformed to queries $q$, keys $k$, and values $v$ as follows:
\begin{equation}
	[q,k,v] = Z[W_{q},W_{k},W_{v}]
\end{equation}
where $W_{q},W_{k}\in \mathbb{R}^{D\times d_{q}},W_{v}\in \mathbb{R}^{D\times d_{v}}$. Then, the output of each encoder block corresponding to the steering token is calculated by:

\begin{equation}
	b^{2} = \sum_{i=1}^{n}a_{2,i}^{'}v^{i}
\end{equation}
where the attention weights are updated by softmax operation, $a_{2,i}^{'} = softmax(q^{2}\bullet k^{i}/\sqrt{d_{q}})$, with $\bullet$ referring dot product operation.

\subsection{Heterogeneous domains-steering supervision (HDSS)}
The hard one-hot label leads to homogenous supervisory signals that hinder the further improvement of FER performance.
Recently, the progress in CLIP \cite{radford2021learning} allows the thriving development of multi-modal models (image and text) in many fields \cite{conde2021clip, wang2022clip} and inspires our work.
Our main motivation is to make image features also have text-space semantic correlations by supervising the similarity between image features and text features, thereby easing the issue of annotation ambiguity.
In this section, we describe in detail how to realize heterogeneous domains-steering supervision by calculating image-text similarity.
The pipeline of HDSS is shown in Fig \ref{fig1}.

Given a dataset contains $M$ expressions, a vector of text label can be generated from one-hot labels, denoted as $t = \{t_{1}; t_{2}; \ldots; t_{M}\}$.
For example, the text label for RAF-DB dataset can be denoted as: "this is a face image of \{expression\}", where \{expression\} belongs to [surprise; fear; disgust; happy; sad; angry; neutral].
The image feature corresponding to steering token is denoted as $I_{S} \in \mathbb{R}^{1\times D}$, and the text features from text encoder are denoted as $T = \{T_{1}; T_{2}; \ldots; T_{M}\}$, $T \in \mathbb{R}^{M\times D}$.
Then, cosine similarity between an image feature $I_{S}$ and a text features $T$ is calculated by $S_{i-t} = I_{S}\times T^{\mathrm{T}}$, where $S_{i-t} \in \mathbb{R}^{1\times M}$. The result after softmax operation is directly used for calculating cross-entropy loss.
\begin{equation}
	P_{S_{i-t}^{i}} = \frac{\exp(S_{i-t}^{i})}{\sum_{j}\exp(S_{i-t}^{j})},
	\mathcal{L}_{text} = - \sum_{i}y_{i}log(P_{S_{i-t}^{i}})
\end{equation}

In addition, the image feature corresponding to class token is denoted as $I_{C} \in \mathbb{R}^{1\times D}$, which is firstly linearly projected to $S_{i-l} \in \mathbb{R}^{1\times M}$ by a fully-connected layer: $S_{i-l} = proj(I_{C})$. The result after softmax operation is also used for calculating another cross-entropy loss.
\begin{equation}
	P_{S_{i-l}^{i}} = \frac{\exp(S_{i-l}^{i})}{\sum_{j}\exp(S_{i-l}^{j})},
	\mathcal{L}_{image} = - \sum_{i}y_{i}log(P_{S_{i-l}^{i}})
\end{equation}
where $y_{i}$ is the ground truth.

Finally, the joint loss function is defined as $\mathcal{L} = \mathcal{L}_{text} + \mathcal{L}_{image} \label{equat1}$.

%-------------------------------------------------------------------------
\section{Experiments}
\textbf{Datasets:}
RAF-DB \cite{li2019reliable}
is one of the most widely used large-scale real-world FER database.
The single-label subset (six basic expression and neutral), consisting of 12,771 training images and 3,068 test images, is used in the experiments.
The overall accuracy is reported on the test set.
FERPlus \cite{BarsoumICMI2016}
extends from FER 2013 \cite{goodfellow2013challenges} and contains eight expression categories (six basic expressions, plus neutral and contempt).
It consists of 28,709 training, 3,589 validation, and 3,589 test images.
The overall accuracy is reported on the test set.
SFEW \cite{dhall2011static}
dataset is created by selecting static frames from the AFEW database.
SFEW 2.0 is the most commonly used version and contains 958 training samples, 436 validation samples, and 372 testing samples.
Each image is assigned to one of seven expression categories, including neutral and the six basic expressions.
The results on validation sets is reported because the labels of the testing set are not released.

\textbf{Implementation details:}
For RAF-DB and SFEW 2.0 datasets, original images are used in all experiments.
For FERPlus dataset, we use MTCNN \cite{zhang2016joint} to detect, align, and crop images.
During training, data augmentation is employed on-the-fly and includes random grayscale, random horizontal flip, and random erasing to avoid over-fitting.
At test time, no data augmentation is used.
All images are resized to $112 \times 112$ pixels before they are fed into the network.
Our model is optimized via the Stochastic Gradient Descent (SGD) optimizer with a mini-batch size of 16.
The learning rate is initialized to 0.0001 and is decayed by a factor of 10 every 30 epochs.
Our experiment is conducted on a GeForce RTX2060 GPU, with PyTorch 1.7.1, Python 3.8.13, and Windows 10.

For RAF-DB and FERPlus datasets, the pre-trained IR-50 \cite{deng2019arcface} on Ms-Celeb-1M \cite{guo2016ms} is adopted as a feature extractor, which is consistent with TranFER \cite{xue2021transfer}.
For SFEW 2.0 dataset, we pre-train FER-former on RAF-DB dataset and then fine-tune it on SFEW 2.0 dataset, which is consistent with IPD-FER \cite{jiang2022disentangling}.
Regarding our scratch-trained Transformer encoder, the depth is 16, the embedding dimension is 256, the number of heads is 4, and the mlp ratio is 4.

\subsection{Comparison with state-of-the-art methods}
We compare our results to several state-of-the-art methods on RAF-DB, FERPlus, and SFEW 2.0 datasets.

\textbf{Comparison on RAF-DB dataset}  between our result and several state-of-the-art methods is shown in Table \ref{tab1}.
RAF-DB dataset is one of the most widely used large-scale real-world FER datasets because it facilitates fair comparisons, in which all images are cropped and do not require any additional preprocessing.
Results show that our FaceFormer achieves state-of-the-art performance compared to all other methods, including FER with unconstrained variations (RAN \cite{wang2020region}, MA-Net \cite{zhao2021learning}, IPD-FER \cite{jiang2022disentangling}), and FER with annotation ambiguity (SCN \cite{wang2020suppressing}, DMUE \cite{she2021dive}, KTN \cite{li2021adaptively}, EfficientFace \cite{zhao2021robust}, SPLDL\cite{shao2022self}, EASE\cite{wang2022ease}).
In particular, when compared to TransFER \cite{xue2021transfer}, the previous best achieved by combining CNN and ViT, FER-former lowers the error rate from 9.09\% to 8.7\%, a 4.3\% improvement.
\begin{table}[h]
	\setlength{\belowcaptionskip}{0.2cm}
	\renewcommand\arraystretch{0.8}
	\caption{Performance comparison with the state-of-the-art methods on RAF-DB dataset.}
	\label{tab1}
	\centering
    \setlength{\tabcolsep}{5.5mm}{
	\begin{tabular}{ccc}
		\toprule
		Methods     &  Years &  Acc.(\%)   \\
		\midrule
		RAN \cite{wang2020region} &  2020  &  86.90  \\
		SCN \cite{wang2020suppressing}&   2020 &  88.14 \\
		DLN \cite{zhang2021learning}&  2021 &  86.40  \\
		%DACL \cite{farzaneh2021facial}& 2021 &  87.78 \\
		KTN \cite{li2021adaptively}&  2021 &  88.07 \\
		MA-Net \cite{zhao2021learning}&  2021 &  88.40 \\
		DMUE \cite{she2021dive}&  2021 &  89.42 \\
		EfficientFace \cite{zhao2021robust}& 2021 & 88.36 \\
		TransFER \cite{xue2021transfer}&  2021  & 90.91 \\
		%HRL \cite{han2022devil}&  2022 & 87.77 \\
		IPD-FER \cite{jiang2022disentangling}&  2022 & 88.89 \\
		CRS-CONT \cite{li2022crs} &  2022 & 88.07 \\
		SPLDL\cite{shao2022self} & 2022 & 89.08 \\
		EASE\cite{wang2022ease} & 2022 & 89.56 \\
		FER-former (Ours)&   2023     &    \textbf{91.30}     \\
		\bottomrule
	\end{tabular}}
\end{table}

\textbf{Comparison on FERPlus dataset}  is shown in Table \ref{tab2}.
It can been seen that our FER-former achieved best FER performance compared to other methods, including FER with unconstrained variations (RAN \cite{wang2020region}, IPD-FER \cite{jiang2022disentangling}), and FER with annotation ambiguity (SCN \cite{wang2020suppressing}, DMUE \cite{she2021dive}, KTN \cite{li2021adaptively}, EASE \cite{wang2022ease}).
In particular, FER-former still shows superiority over previous best results reported in TransFER \cite{xue2021transfer}.
\begin{table}[h]
	\setlength{\belowcaptionskip}{0.2cm}
	\renewcommand\arraystretch{0.8}
	\caption{Performance comparison with the state-of-the-art methods on FERPlus dataset.}
	\label{tab2}
	\centering
    \setlength{\tabcolsep}{5.5mm}{
	\begin{tabular}{ccc}
		\toprule
		Methods     & Years &  Acc.(\%) \\
		\midrule
		RAN \cite{wang2020region}& 2020  &  88.55  \\
		SCN \cite{wang2020suppressing}& 2020  & 89.35 \\
		DMUE \cite{she2021dive}&  2021   & 89.51 \\
		KTN \cite{li2021adaptively}&2021 & 90.49   \\
		TransFER \cite{xue2021transfer}& 2021 & 90.83 \\
		IPD-FER \cite{jiang2022disentangling}& 2022 & 88.42 \\
		EASE \cite{wang2022ease}& 2022 & 90.26 \\
		FER-former (Ours) & 2023 & \textbf{90.96} \\
		\bottomrule
	\end{tabular}}
\end{table}

\textbf{Comparison on SFEW 2.0 dataset} is shown in Table \ref{tab3}.
Due to the limited images, we pre-trained FER-former on RAF-DB dataset and fine-tune it on SFEW 2.0 dataset because they have the same expression categories, which is in consistent with IPD-FER \cite{jiang2022disentangling}.
It can be observed that CRS-CONT \cite{li2022crs}, EASE \cite{wang2022ease} and FER-former are the only three models to achieve accuracy over 60\% on this dataset.
FER-former improves FER accuracy by 2.06\% compared to EASE \cite{wang2022ease}, achieving state-of-the-art performance.
\begin{table}[h]
	\setlength{\belowcaptionskip}{0.2cm}
	\renewcommand\arraystretch{0.8}
	\caption{Performance comparison with the state-of-the-art methods on SFEW 2.0 dataset.}
	\label{tab3}
	\centering
    \setlength{\tabcolsep}{5.5mm}{
	\begin{tabular}{ccc}
		\toprule
		Methods    & Years  & Acc.(\%)  \\
		\midrule
		Incept-ResV1 \cite{acharya2018covariance} & 2018 & 51.90 \\
		Island loss \cite{cai2018island} & 2018 & 52.52 \\
		LTNet \cite{zeng2018facial} & 2018 & 58.29 \\
		RAN \cite{wang2020region}     & 2020 & 56.40 \\
		MA-Net \cite{zhao2021learning}     &  2021  & 59.40 \\
		DMUE \cite{she2021dive}     &  2021   & 58.34 \\
		IPD-FER \cite{jiang2022disentangling} & 2022 & 58.43 \\
		CRS-CONT \cite{li2022crs} & 2022 & 60.09 \\
		EASE \cite{wang2022ease}& 2022 & 60.12 \\
		FER-former (Ours)     &   2023  &  \textbf{62.18} \\
		\bottomrule
	\end{tabular}}
	\label{tab:table}
\end{table}

\subsection{Confusion matrices}
\begin{figure}[h]
	\centering
	\includegraphics[width=0.1465\textwidth]{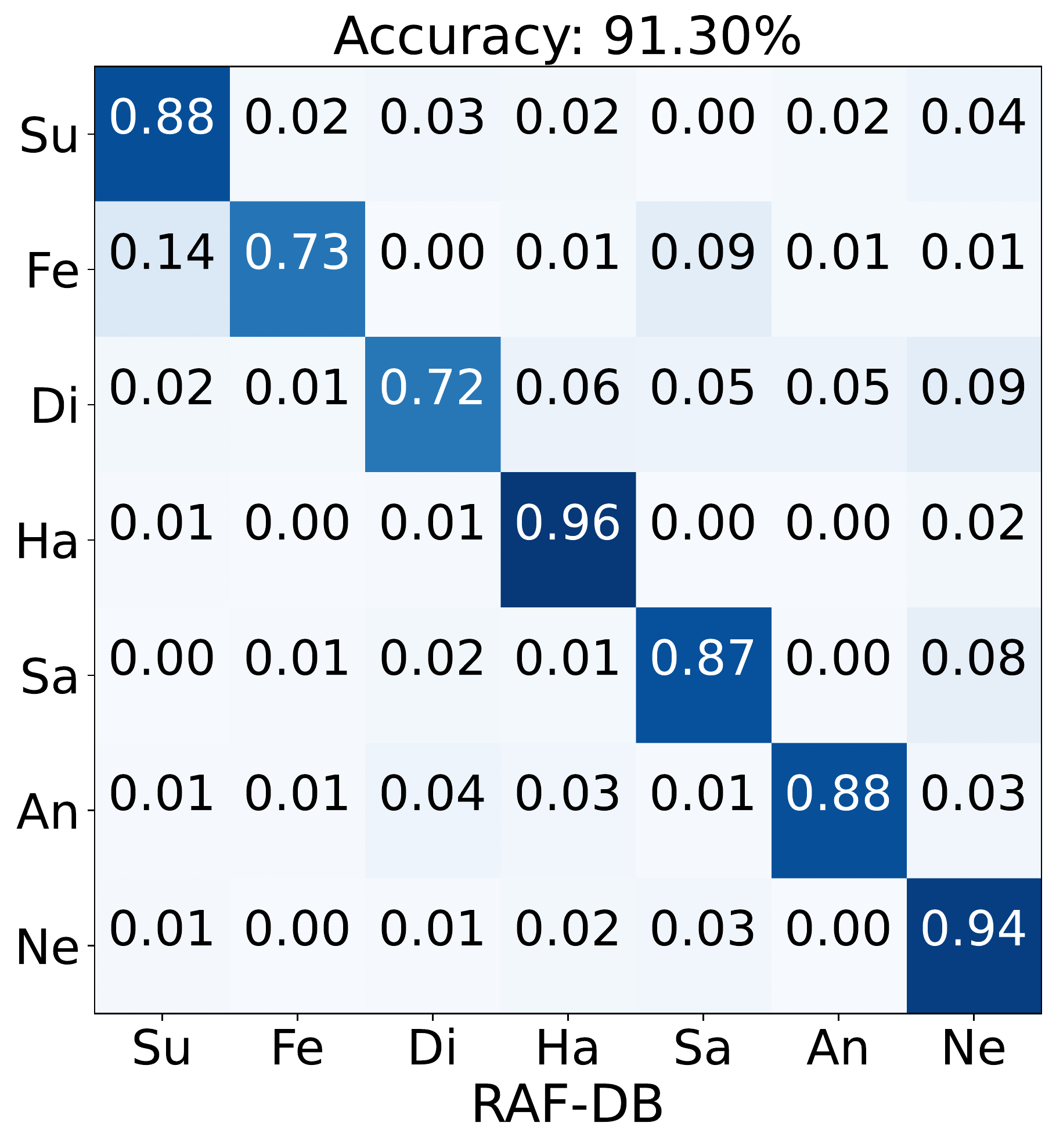}
    \includegraphics[width=0.1465\textwidth]{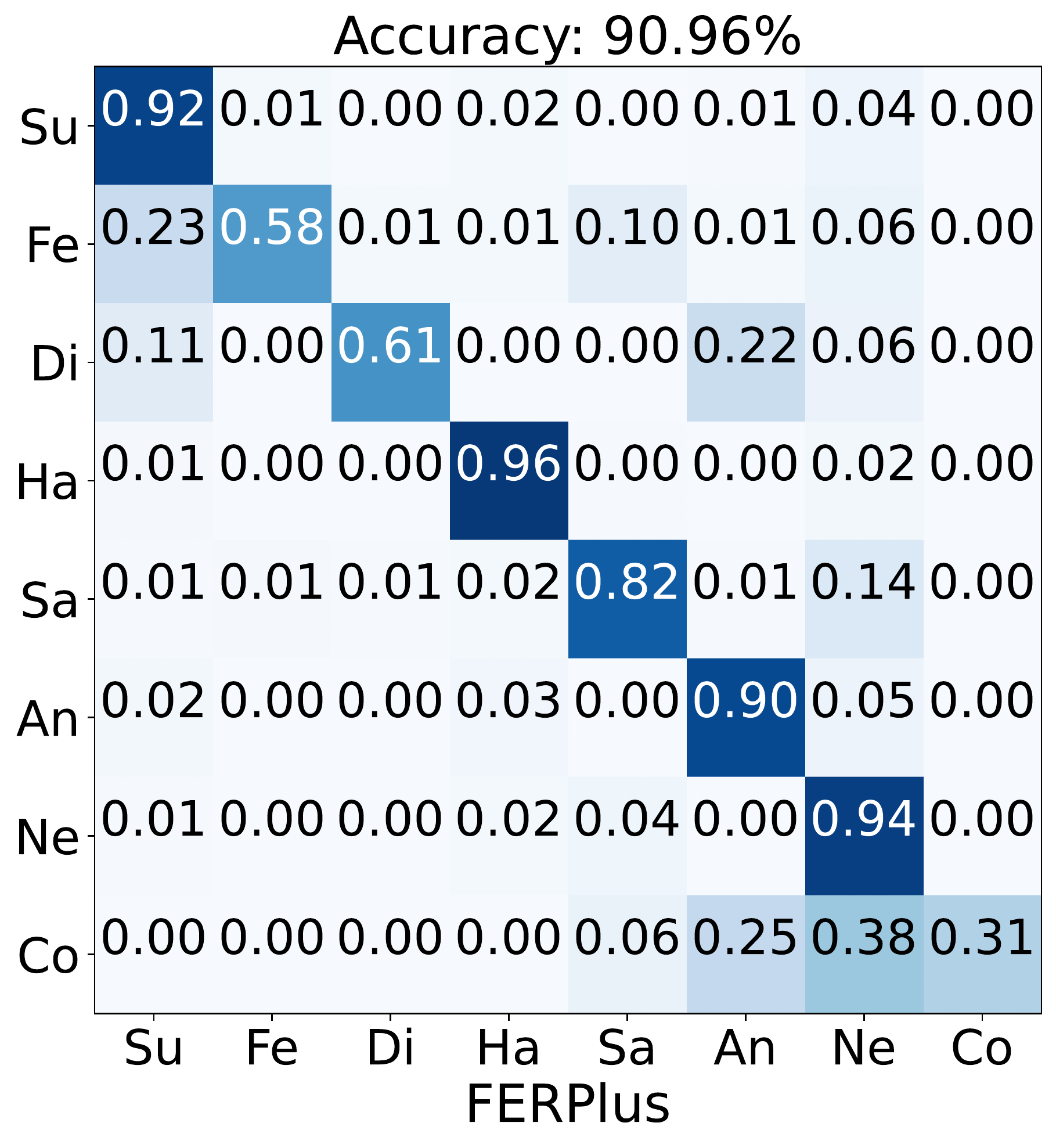}
    \includegraphics[width=0.175\textwidth]{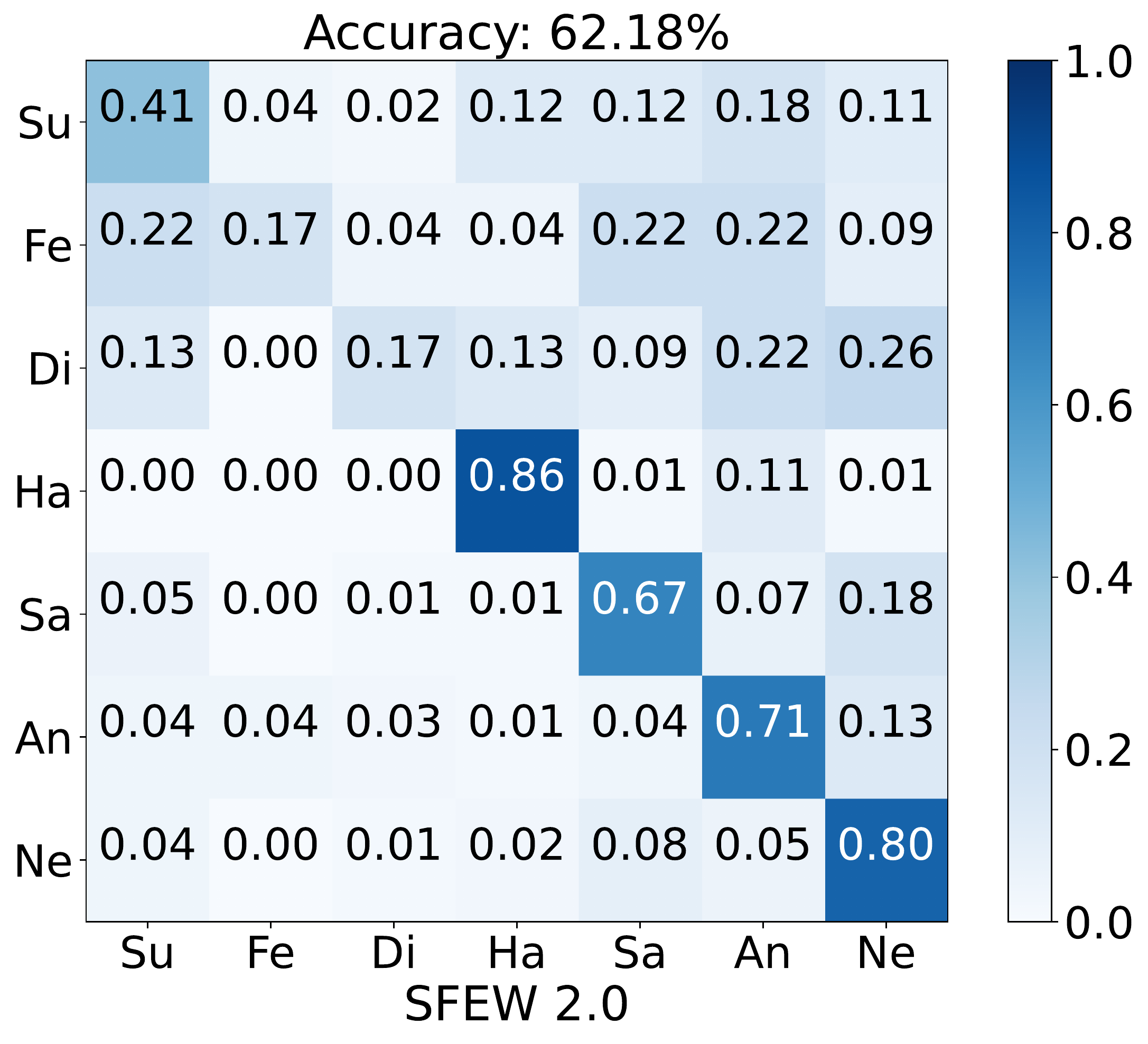}
	\caption{Confusion matrices of FER-former on RAF-DB, FERPlus, and SFEW 2.0 datasets, where y-axis and x-axis represent ground-truth and predicted labels, respectively (Su: surprise, Fe: fear, Di: disgust, Ha: happy, Sa: sad, An: anger, Ne: neutral, Co: contempt).}
	\label{fig3}
\end{figure}

The confusion matrices of our FER-former on the three datasets are shown in Fig \ref{fig3}.
It can be observed that the FER accuracy of ``fear", ``disgust" and ``contempt" is noticeably lower than that of other expressions.
In particular, ``fear" is likely to confused with ``surprise" and ``sad" with ``neutral" on all three datasets, indicating the proximity of these expressions in the semantic space.
The accuracy of ``happy" and ``neutral" are higher than that of other expressions, suggesting that they are less ambiguous and easier to distinguish from other expressions.

\begin{figure}[h]
	\centering
	\includegraphics[width=0.48\textwidth]{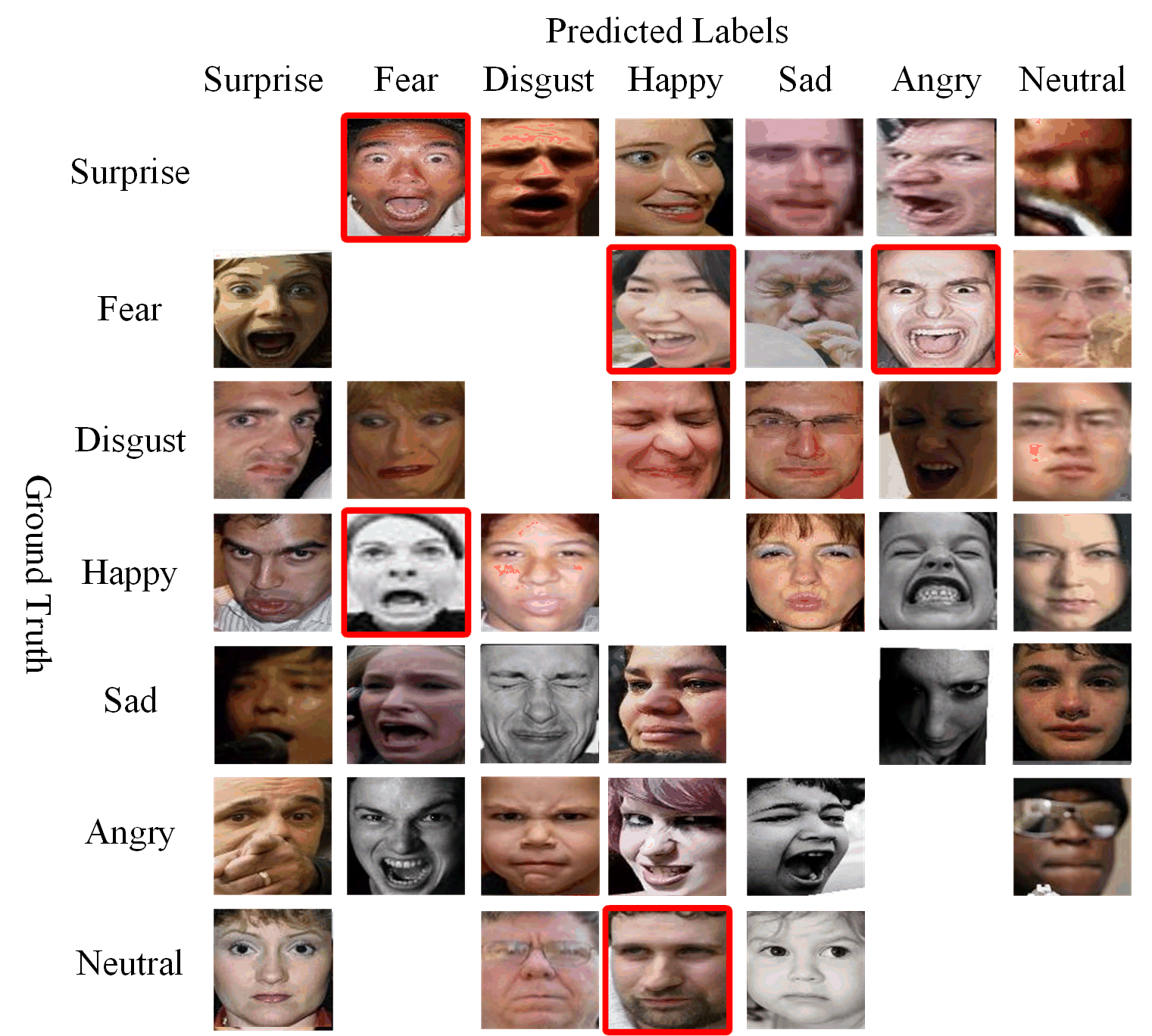}
	\caption{Some mislabelled examples on RAF-DB test set, where y-axis and x-axis represent ground-truth and predicted labels, respectively. The empty spaces indicate that there is no mislabelled sample in the corresponding categories. The samples highlighted by red boxes are those that we consider to be highly ambiguous.}
	\label{fig5}
\end{figure}

To analyze when FER-former mislabels impressions, selected images from different confusion categories are shown in Fig \ref{fig5}.
It can be observed that ambiguous samples are widespread in all expression categories. %, including ambiguous facial expressions and low-quality facial images.
Close inspection on the highlighted mislabelled samples shows that the labels produced by FER-former are sometime arguably more convincing than the ground truth.

\subsection{Ablation study}
\textbf{Evaluation of MGEI, HDSS on RAF-DB and FERPlus datasets.}
To validate the effectiveness of MGEI and HDSS modules in our FER-former, an ablation study was conducted on RAF-DB and FERPlus datasets, as shown in Table \ref{tab:table4}.
The baseline in the first row does not use MGEI nor HDSS.
The extracted features from stem CNN are sliced with local size $2 \times 2 $ along spatial dimension.
Compared to the baseline, the accuracy is improved by 0.65\% and 0.32\% on the two datasets when adding MGEI module and by 1.01\% and 0.64\% when adding HDSS. The combination of MGEI and HDSS can noticeably boost the performance of FER in the wild and achieves the best results.
\begin{table}[h]
	\setlength{\belowcaptionskip}{0.2cm}
	\renewcommand\arraystretch{0.8}
	\caption{Evaluation (\%) of MGEI, HDSS on RAF-DB and FERPlus datasets.}
	\centering
    \setlength{\tabcolsep}{4.5mm}{
	\begin{tabular}{cc|cc}
		\toprule
		MGEI    &  HDSS  &  RAF-DB  & FERPlus \\
		\midrule
		&    &  89.80   &  90.07    \\
		$\surd$ &    &  90.45  &   90.39    \\
		& $\surd$  &  90.81  &  90.71   \\
		$\surd$ & $\surd$  &  \textbf{91.30}  &  \textbf{90.96}   \\
		\bottomrule
	\end{tabular}}
	\label{tab:table4}
\end{table}

\begin{figure}[h]
	\centering
	\includegraphics[width=0.48\textwidth]{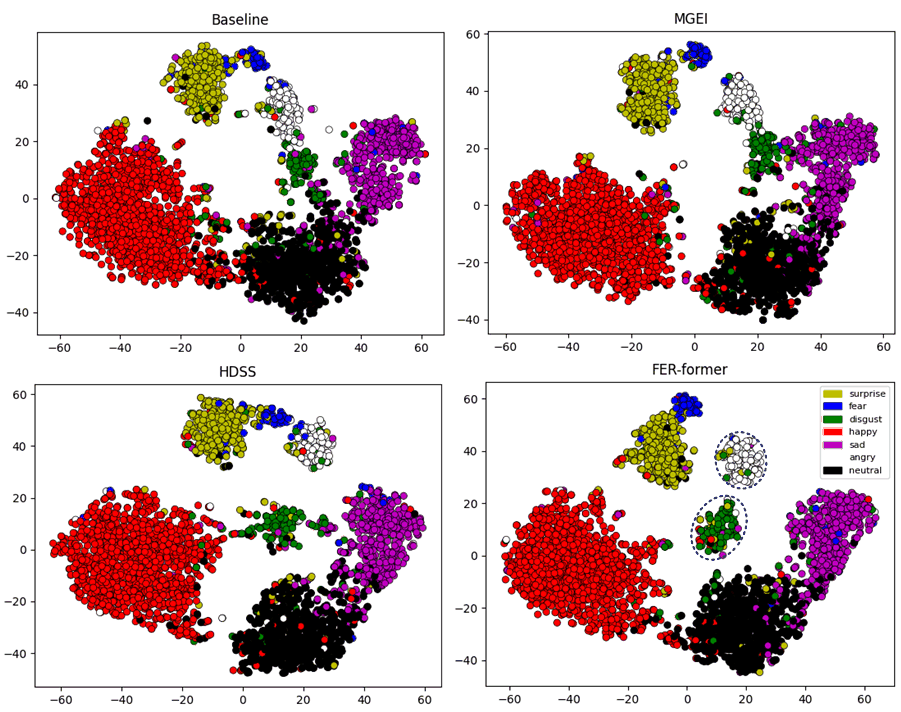}
	\caption{Visualization (t-SNE \cite{van2008visualizing}) of high dimension features generated under different settings on the RAF-DB dataset. Each colour-coded dot represents a test image with the corresponding labelled expression.} %It can be observed that FER-former is capable of increasing inter-class distance, as well as decreasing inner-class distance.}
	\label{fig4}
\end{figure}

To more intuitively illustrate the effectiveness of MGEI and HDSS modules, the high dimensional image features corresponding to the steering token from the last encoder is visualized by t-SNE \cite{van2008visualizing}.
As shown in Fig \ref{fig4}, both MGEI and HDSS are effective in maximizing the margins between expression classes.
With both, FER-former can effectively separate ``angry" and ``disgust" from other expressions.
In addition, we can observe that ``fear" is always close to ``surprise" and ``sad" is always close to ``neutral" in high dimension space, which aligns well with the above observations.

%\begin{figure}[h]
%	\centering
%	\includegraphics[width=0.45\textwidth]{./img/anotation_ambiguous.png}
%	\caption{Annotation ambiguous-robust FER. Example results are shown on the RAF-DB test set. The labels above each expression indicate the prediction results of FER-former (red) and MGEI (with text supervision removed) , respectively.}
%	\label{fig4}
%\end{figure}
\textbf{Evaluation of different text contents on RAF-DB and FERPlus datasets.}
Text content is introduced to mitigate annotation ambiguity by calculating cosine similarity between image features and text features, which plays an important role in our proposed FER-former.
In order to evaluate the robustness of FER-former to different text contents, comparative experiments of five cases were conducted, including case 1: without using text, case 2: using expression name as a single word, case 3: using a phrase in the format of ``a face image of \{\}'', case 4: using a full active sentence ``this is a face image of \{\}'', and case 5: using a passive sentence ``a/an \{\} expression is shown in the image''.
The detailed results are reported in Table \ref{tab:table5}.

\begin{table}[h]
	\caption{Evaluation of different text contents on RAF-DB and FERPlus datasets.}
	\centering
	\setlength{\belowcaptionskip}{0.2cm}
	\renewcommand\arraystretch{0.8}
    \setlength{\tabcolsep}{5.5mm}{
		\begin{tabular}{clc}
			\toprule
			Datasets  & Texts contents  &  Acc.(\%) \\
			\midrule
			\multirow{5}*{RAFDB} & without text   &  90.45   \\
			~& single word   &   90.81   \\
			~& phrase   &   \textbf{91.30}   \\
			~& active sentence   &    91.13  \\
			~& passive sentence   &    90.97  \\
			\midrule
			\multirow{5}*{FERPlus} & without text   &  90.39   \\
			~& single word   &  90.61    \\
			~& phrase   &   90.90   \\
			~& active sentence   &    \textbf{90.96}  \\
			~& passive sentence   &  90.87   \\
			\bottomrule
	\end{tabular}}
	\label{tab:table5}
\end{table}
On RAF-DB dataset, the accuracy of four cases with text supervision surpasses that of without text, which shows the positive effects of text contents.
In addition, single word performs worst among the several text cases, and active sentence outperforms passive sentence.
Our hypothesis is that single expression words lack context whereas the passive sentences are too complicated.
The results indicate that contextual semantics play an important role in the text encoder.
More importantly, the performance of the phrase achieves state-of-the-art results with a 0.85\% improvement compared to case 1, which shows the superiority of our FER-former.
For FERPlus dataset, most of the conclusions are consistent with the above.
It is worth emphasizing that the active sentence rather than the phrase performs best on FERPlus dataset.
We consider that both active sentence and phrase are capable of achieving state-of-the-art results, and the difference in results depends on different model initialization parameters.

\textbf{Evaluation of patch size on RAF-DB and FERPlus datasets.}
To enable our ER-former to overcome the issues of occlusion and pose variances, the extracted features are split into sequences of patches with varied resolutions along spatial dimension.
The above results have shown that MGEI has a great impact on the performance of FER.
Here we aim to investigate the impact of local patches of different resolutions, as shown in Table \ref{tab:table6}.
It can be seen that each single division method behaves differently on different datastes.
For example, $4*4$ performs better on RAF-DB dataset and $2*2$ performs better on FERPlus dataset.
We believe this is due to the huge differences between the original images of different datasets.
Therefore, we propose to integrate multi-granularity embeddings to achieve resolution-robust feature division and occlusion and pose-robust feature learning.
The current results show that the integration of (2,4,6,12) can perform best on the both datasets, further indicating the positive effect of MGEI.

\begin{table}[h]
	\setlength{\belowcaptionskip}{0.2cm}
	\renewcommand\arraystretch{0.8}
	\caption{Evaluation of patch resolution on RAF-DB and FERPlus datasets.}
	\centering
	\setlength{\tabcolsep}{1mm}{
		\begin{tabular}{ccccccc}
			\toprule
			Datasets & 2*2 & 3*3 & 4*4 & 6*6 & (2,4,6) & (2,4,6,12)\\
			\midrule
			RAFDB &  90.81  &  90.38  &  90.91   &  90.48   &   90.81    &   \textbf{91.30}  \\
			FERPlus &  90.71  &  90.39  &  90.10   &  90.42  &   90.45   &   \textbf{90.96}  \\
			\bottomrule
	\end{tabular}}
	\label{tab:table6}
\end{table}

%-------------------------------------------------------------------------
\section{Conclusion}
This paper presents a multifarious supervision-steering Transformer for FER in the wild, namely FER-former, which features multi-granularity embedding integration, hybrid self-attention scheme and multifarious supervisory signals from heterogeneous domains.
In general, a pre-trained prevailing CNNs (features pyramid) and a scratch-trained Transformer (global receptive fields) are combined to take full advantage of their merits.
In particular, a FER-specific encoder containing a hybrid self-attention scheme is well devised to characterize ambiguity-robust tokens involving rich one-hot and text semantic.
More importantly, a heterogeneous domains-steering supervision module is proposed to mitigate annotation ambiguity by supervising the cosine similarity between image features and text features, which plays an important role in our FER-former.
Finally, on top of the collaboration of heterogeneous token heads, fine-grained global interrelationships with diverse semantic cues among tokens are enriched and modeled.
Results on several benchmarks show the superiority of our FER-former over other state-of-the-art methods.

%%%%%%%%% REFERENCES
{\small
\bibliographystyle{IEEEtran}
\bibliography{references}

% Generated by IEEEtran.bst, version: 1.13 (2008/09/30)
\begin{thebibliography}{10}
\providecommand{\url}[1]{#1}
\csname url@samestyle\endcsname
\providecommand{\newblock}{\relax}
\providecommand{\bibinfo}[2]{#2}
\providecommand{\BIBentrySTDinterwordspacing}{\spaceskip=0pt\relax}
\providecommand{\BIBentryALTinterwordstretchfactor}{4}
\providecommand{\BIBentryALTinterwordspacing}{\spaceskip=\fontdimen2\font plus
\BIBentryALTinterwordstretchfactor\fontdimen3\font minus
  \fontdimen4\font\relax}
\providecommand{\BIBforeignlanguage}[2]{{%
\expandafter\ifx\csname l@#1\endcsname\relax
\typeout{** WARNING: IEEEtran.bst: No hyphenation pattern has been}%
\typeout{** loaded for the language `#1'. Using the pattern for}%
\typeout{** the default language instead.}%
\else
\language=\csname l@#1\endcsname
\fi
#2}}
\providecommand{\BIBdecl}{\relax}
\BIBdecl

\bibitem{li2022dual}
Y.~Li, Y.~Lu, M.~Gong, L.~Liu, and L.~Zhao, ``Dual-channel feature
  disentanglement for identity-invariant facial expression recognition,''
  \emph{Information Sciences}, vol. 608, pp. 410--423, 2022.

\bibitem{yang2020novel}
L.~Yang, Y.~Tian, Y.~Song, N.~Yang, K.~Ma, and L.~Xie, ``A novel feature
  separation model exchange-gan for facial expression recognition,''
  \emph{Knowledge-Based Systems}, vol. 204, p. 106217, 2020.

\bibitem{valstar2010induced}
M.~Valstar, M.~Pantic \emph{et~al.}, ``Induced disgust, happiness and surprise:
  an addition to the mmi facial expression database,'' in \emph{Proc. 3rd
  Intern. Workshop on EMOTION (satellite of LREC): Corpora for Research on
  Emotion and Affect}.\hskip 1em plus 0.5em minus 0.4em\relax Paris, France.,
  2010, p.~65.

\bibitem{lucey2010extended}
P.~Lucey, J.~F. Cohn, T.~Kanade, J.~Saragih, Z.~Ambadar, and I.~Matthews, ``The
  extended cohn-kanade dataset (ck+): A complete dataset for action unit and
  emotion-specified expression,'' in \emph{2010 ieee computer society
  conference on computer vision and pattern recognition-workshops}.\hskip 1em
  plus 0.5em minus 0.4em\relax IEEE, 2010, pp. 94--101.

\bibitem{zhao2011facial}
G.~Zhao, X.~Huang, M.~Taini, S.~Z. Li, and M.~Pietik{\"a}Inen, ``Facial
  expression recognition from near-infrared videos,'' \emph{Image and vision
  computing}, vol.~29, no.~9, pp. 607--619, 2011.

\bibitem{zhou2022contextual}
Q.~Zhou, X.~Wu, S.~Zhang, B.~Kang, Z.~Ge, and L.~J. Latecki, ``Contextual
  ensemble network for semantic segmentation,'' \emph{Pattern Recognition},
  vol. 122, p. 108290, 2022.

\bibitem{wang2021end}
J.~Wang, L.~Song, Z.~Li, H.~Sun, J.~Sun, and N.~Zheng, ``End-to-end object
  detection with fully convolutional network,'' in \emph{Proceedings of the
  IEEE/CVF conference on computer vision and pattern recognition}, 2021, pp.
  15\,849--15\,858.

\bibitem{sun2021supervised}
H.~Sun, X.~Zheng, and X.~Lu, ``A supervised segmentation network for
  hyperspectral image classification,'' \emph{IEEE Transactions on Image
  Processing}, vol.~30, pp. 2810--2825, 2021.

\bibitem{kahou2013combining}
S.~E. Kahou, C.~Pal, X.~Bouthillier, P.~Froumenty, {\c{C}}.~G{\"u}l{\c{c}}ehre,
  R.~Memisevic, P.~Vincent, A.~Courville, Y.~Bengio, R.~C. Ferrari
  \emph{et~al.}, ``Combining modality specific deep neural networks for emotion
  recognition in video,'' in \emph{Proceedings of the 15th ACM on International
  conference on multimodal interaction}, 2013, pp. 543--550.

\bibitem{tang2013deep}
Y.~Tang, ``Deep learning using linear support vector machines,'' \emph{arXiv
  preprint arXiv:1306.0239}, 2013.

\bibitem{wang2020region}
K.~Wang, X.~Peng, J.~Yang, D.~Meng, and Y.~Qiao, ``Region attention networks
  for pose and occlusion robust facial expression recognition,'' \emph{IEEE
  Transactions on Image Processing}, vol.~29, pp. 4057--4069, 2020.

\bibitem{LiZSC19}
Y.~Li, J.~Zeng, S.~Shan, and X.~Chen, ``Occlusion aware facial expression
  recognition using {CNN} with attention mechanism,'' \emph{{IEEE} Trans. Image
  Process.}, vol.~28, no.~5, pp. 2439--2450, 2019.

\bibitem{zhao2021learning}
Z.~Zhao, Q.~Liu, and S.~Wang, ``Learning deep global multi-scale and local
  attention features for facial expression recognition in the wild,''
  \emph{IEEE Transactions on Image Processing}, vol.~30, pp. 6544--6556, 2021.

\bibitem{han2019enhanced}
Y.~Han, B.~Tang, and L.~Deng, ``An enhanced convolutional neural network with
  enlarged receptive fields for fault diagnosis of planetary gearboxes,''
  \emph{Computers in Industry}, vol. 107, pp. 50--58, 2019.

\bibitem{gu2017enlarging}
Y.~Gu, Z.~Zhong, S.~Wu, and Y.~Xu, ``Enlarging effective receptive field of
  convolutional neural networks for better semantic segmentation,'' in
  \emph{2017 4th IAPR Asian Conference on Pattern Recognition (ACPR)}.\hskip
  1em plus 0.5em minus 0.4em\relax IEEE, 2017, pp. 388--393.

\bibitem{vaswani2017attention}
A.~Vaswani, N.~Shazeer, N.~Parmar, J.~Uszkoreit, L.~Jones, A.~N. Gomez,
  {\L}.~Kaiser, and I.~Polosukhin, ``Attention is all you need,''
  \emph{Advances in neural information processing systems}, vol.~30, 2017.

\bibitem{devlin2018bert}
J.~Devlin, M.-W. Chang, K.~Lee, and K.~Toutanova, ``Bert: Pre-training of deep
  bidirectional transformers for language understanding,'' \emph{arXiv preprint
  arXiv:1810.04805}, 2018.

\bibitem{dosovitskiy2020image}
A.~Dosovitskiy, L.~Beyer, A.~Kolesnikov, D.~Weissenborn, X.~Zhai,
  T.~Unterthiner, M.~Dehghani, M.~Minderer, G.~Heigold, S.~Gelly \emph{et~al.},
  ``An image is worth 16x16 words: Transformers for image recognition at
  scale,'' \emph{arXiv preprint arXiv:2010.11929}, 2020.

\bibitem{gu2022multi}
J.~Gu, H.~Kwon, D.~Wang, W.~Ye, M.~Li, Y.-H. Chen, L.~Lai, V.~Chandra, and
  D.~Z. Pan, ``Multi-scale high-resolution vision transformer for semantic
  segmentation,'' in \emph{Proceedings of the IEEE/CVF Conference on Computer
  Vision and Pattern Recognition}, 2022, pp. 12\,094--12\,103.

\bibitem{sun2022spectral}
L.~Sun, G.~Zhao, Y.~Zheng, and Z.~Wu, ``Spectral--spatial feature tokenization
  transformer for hyperspectral image classification,'' \emph{IEEE Transactions
  on Geoscience and Remote Sensing}, vol.~60, pp. 1--14, 2022.

\bibitem{han2022few}
G.~Han, J.~Ma, S.~Huang, L.~Chen, and S.-F. Chang, ``Few-shot object detection
  with fully cross-transformer,'' in \emph{Proceedings of the IEEE/CVF
  Conference on Computer Vision and Pattern Recognition}, 2022, pp. 5321--5330.

\bibitem{tu2022maxim}
Z.~Tu, H.~Talebi, H.~Zhang, F.~Yang, P.~Milanfar, A.~Bovik, and Y.~Li, ``Maxim:
  Multi-axis mlp for image processing,'' in \emph{Proceedings of the IEEE/CVF
  Conference on Computer Vision and Pattern Recognition}, 2022, pp. 5769--5780.

\bibitem{xue2021transfer}
F.~Xue, Q.~Wang, and G.~Guo, ``Transfer: Learning relation-aware facial
  expression representations with transformers,'' in \emph{ICCV}, 2021, pp.
  3601--3610.

\bibitem{chen2021understanding}
Y.~Chen and J.~Joo, ``Understanding and mitigating annotation bias in facial
  expression recognition,'' in \emph{Proceedings of the IEEE/CVF International
  Conference on Computer Vision}, 2021, pp. 14\,980--14\,991.

\bibitem{mao2021label}
S.~Mao, G.~Shi, L.~Jiao, S.~Gou, Y.~Li, L.~Xiong, and B.~Shi, ``Label
  distribution amendment with emotional semantic correlations for facial
  expression recognition,'' \emph{arXiv preprint arXiv:2107.11061}, 2021.

\bibitem{zhao2021robust}
Z.~Zhao, Q.~Liu, and F.~Zhou, ``Robust lightweight facial expression
  recognition network with label distribution training,'' in \emph{AAAI},
  vol.~35, no.~4, 2021, pp. 3510--3519.

\bibitem{chen2020label}
S.~Chen, J.~Wang, Y.~Chen, Z.~Shi, X.~Geng, and Y.~Rui, ``Label distribution
  learning on auxiliary label space graphs for facial expression recognition,''
  in \emph{Proceedings of the IEEE/CVF conference on computer vision and
  pattern recognition}, 2020, pp. 13\,984--13\,993.

\bibitem{she2021dive}
J.~She, Y.~Hu, H.~Shi, J.~Wang, Q.~Shen, and T.~Mei, ``Dive into ambiguity:
  Latent distribution mining and pairwise uncertainty estimation for facial
  expression recognition,'' in \emph{CVPR}, 2021, pp. 6248--6257.

\bibitem{shao2022self}
J.~Shao, Z.~Wu, Y.~Luo, S.~Huang, X.~Pu, and Y.~Ren, ``Self-paced label
  distribution learning for in-the-wild facial expression recognition,'' in
  \emph{Proceedings of the 30th ACM International Conference on Multimedia},
  2022, pp. 161--169.

\bibitem{wang2020suppressing}
K.~Wang, X.~Peng, J.~Yang, S.~Lu, and Y.~Qiao, ``Suppressing uncertainties for
  large-scale facial expression recognition,'' in \emph{CVPR}, 2020, pp.
  6897--6906.

\bibitem{Li2022towards}
H.~Li, N.~Wang, X.~Yang, X.~Wang, and X.~Gao, ``Towards semi-supervised deep
  facial expression recognition with an adaptive confidence margin,'' in
  \emph{CVPR}, 2022.

\bibitem{wang2022ease}
L.~Wang, G.~Jia, N.~Jiang, H.~Wu, and J.~Yang, ``Ease: Robust facial expression
  recognition via emotion ambiguity-sensitive cooperative networks,'' in
  \emph{Proceedings of the 30th ACM International Conference on Multimedia},
  2022, pp. 218--227.

\bibitem{radford2021learning}
A.~Radford, J.~W. Kim, C.~Hallacy, A.~Ramesh, G.~Goh, S.~Agarwal, G.~Sastry,
  A.~Askell, P.~Mishkin, J.~Clark \emph{et~al.}, ``Learning transferable visual
  models from natural language supervision,'' in \emph{International Conference
  on Machine Learning}.\hskip 1em plus 0.5em minus 0.4em\relax PMLR, 2021, pp.
  8748--8763.

\bibitem{conde2021clip}
M.~V. Conde and K.~Turgutlu, ``Clip-art: contrastive pre-training for
  fine-grained art classification,'' in \emph{Proceedings of the IEEE/CVF
  Conference on Computer Vision and Pattern Recognition}, 2021, pp. 3956--3960.

\bibitem{wang2022clip}
C.~Wang, M.~Chai, M.~He, D.~Chen, and J.~Liao, ``Clip-nerf: Text-and-image
  driven manipulation of neural radiance fields,'' in \emph{Proceedings of the
  IEEE/CVF Conference on Computer Vision and Pattern Recognition}, 2022, pp.
  3835--3844.

\bibitem{pan2019occluded}
B.~Pan, S.~Wang, and B.~Xia, ``Occluded facial expression recognition enhanced
  through privileged information,'' in \emph{Proceedings of the 27th ACM
  international conference on multimedia}, 2019, pp. 566--573.

\bibitem{xia2020occluded}
B.~Xia and S.~Wang, ``Occluded facial expression recognition with step-wise
  assistance from unpaired non-occluded images,'' in \emph{Proceedings of the
  28th ACM International Conference on Multimedia}, 2020, pp. 2927--2935.

\bibitem{zhang2020geometry}
F.~Zhang, T.~Zhang, Q.~Mao, and C.~Xu, ``Geometry guided pose-invariant facial
  expression recognition,'' \emph{IEEE Transactions on Image Processing},
  vol.~29, pp. 4445--4460, 2020.

\bibitem{guo2022cmt}
J.~Guo, K.~Han, H.~Wu, Y.~Tang, X.~Chen, Y.~Wang, and C.~Xu, ``Cmt:
  Convolutional neural networks meet vision transformers,'' in
  \emph{Proceedings of the IEEE/CVF Conference on Computer Vision and Pattern
  Recognition}, 2022, pp. 12\,175--12\,185.

\bibitem{yuan2021incorporating}
K.~Yuan, S.~Guo, Z.~Liu, A.~Zhou, F.~Yu, and W.~Wu, ``Incorporating convolution
  designs into visual transformers,'' in \emph{Proceedings of the IEEE/CVF
  International Conference on Computer Vision}, 2021, pp. 579--588.

\bibitem{wu2021cvt}
H.~Wu, B.~Xiao, N.~Codella, M.~Liu, X.~Dai, L.~Yuan, and L.~Zhang, ``Cvt:
  Introducing convolutions to vision transformers,'' in \emph{Proceedings of
  the IEEE/CVF International Conference on Computer Vision}, 2021, pp. 22--31.

\bibitem{graham2021levit}
B.~Graham, A.~El-Nouby, H.~Touvron, P.~Stock, A.~Joulin, H.~J{\'e}gou, and
  M.~Douze, ``Levit: a vision transformer in convnet's clothing for faster
  inference,'' in \emph{Proceedings of the IEEE/CVF international conference on
  computer vision}, 2021, pp. 12\,259--12\,269.

\bibitem{srinivas2021bottleneck}
A.~Srinivas, T.-Y. Lin, N.~Parmar, J.~Shlens, P.~Abbeel, and A.~Vaswani,
  ``Bottleneck transformers for visual recognition,'' in \emph{Proceedings of
  the IEEE/CVF conference on computer vision and pattern recognition}, 2021,
  pp. 16\,519--16\,529.

\bibitem{li2021align}
J.~Li, R.~Selvaraju, A.~Gotmare, S.~Joty, C.~Xiong, and S.~C.~H. Hoi, ``Align
  before fuse: Vision and language representation learning with momentum
  distillation,'' \emph{Advances in neural information processing systems},
  vol.~34, pp. 9694--9705, 2021.

\bibitem{singh2022flava}
A.~Singh, R.~Hu, V.~Goswami, G.~Couairon, W.~Galuba, M.~Rohrbach, and D.~Kiela,
  ``Flava: A foundational language and vision alignment model,'' in
  \emph{Proceedings of the IEEE/CVF Conference on Computer Vision and Pattern
  Recognition}, 2022, pp. 15\,638--15\,650.

\bibitem{wang2021simvlm}
Z.~Wang, J.~Yu, A.~W. Yu, Z.~Dai, Y.~Tsvetkov, and Y.~Cao, ``Simvlm: Simple
  visual language model pretraining with weak supervision,'' \emph{arXiv
  preprint arXiv:2108.10904}, 2021.

\bibitem{hu2021unit}
R.~Hu and A.~Singh, ``Unit: Multimodal multitask learning with a unified
  transformer,'' in \emph{Proceedings of the IEEE/CVF International Conference
  on Computer Vision}, 2021, pp. 1439--1449.

\bibitem{kim2021vilt}
W.~Kim, B.~Son, and I.~Kim, ``Vilt: Vision-and-language transformer without
  convolution or region supervision,'' in \emph{International Conference on
  Machine Learning}.\hskip 1em plus 0.5em minus 0.4em\relax PMLR, 2021, pp.
  5583--5594.

\bibitem{zhang2021vinvl}
P.~Zhang, X.~Li, X.~Hu, J.~Yang, L.~Zhang, L.~Wang, Y.~Choi, and J.~Gao,
  ``Vinvl: Revisiting visual representations in vision-language models,'' in
  \emph{Proceedings of the IEEE/CVF Conference on Computer Vision and Pattern
  Recognition}, 2021, pp. 5579--5588.

\bibitem{jia2021scaling}
C.~Jia, Y.~Yang, Y.~Xia, Y.-T. Chen, Z.~Parekh, H.~Pham, Q.~Le, Y.-H. Sung,
  Z.~Li, and T.~Duerig, ``Scaling up visual and vision-language representation
  learning with noisy text supervision,'' in \emph{International Conference on
  Machine Learning}.\hskip 1em plus 0.5em minus 0.4em\relax PMLR, 2021, pp.
  4904--4916.

\bibitem{deng2019arcface}
J.~Deng, J.~Guo, N.~Xue, and S.~Zafeiriou, ``Arcface: Additive angular margin
  loss for deep face recognition,'' in \emph{Proceedings of the IEEE/CVF
  conference on computer vision and pattern recognition}, 2019, pp. 4690--4699.

\bibitem{guo2016ms}
Y.~Guo, L.~Zhang, Y.~Hu, X.~He, and J.~Gao, ``Ms-celeb-1m: A dataset and
  benchmark for large-scale face recognition,'' in \emph{European conference on
  computer vision}.\hskip 1em plus 0.5em minus 0.4em\relax Springer, 2016, pp.
  87--102.

\bibitem{li2019reliable}
S.~Li and W.~Deng, ``Reliable crowdsourcing and deep locality-preserving
  learning for unconstrained facial expression recognition,'' \emph{IEEE
  Transactions on Image Processing}, vol.~28, no.~1, pp. 356--370, 2019.

\bibitem{BarsoumICMI2016}
E.~Barsoum, C.~Zhang, C.~Canton~Ferrer, and Z.~Zhang, ``Training deep networks
  for facial expression recognition with crowd-sourced label distribution,'' in
  \emph{ACM International Conference on Multimodal Interaction (ICMI)}, 2016.

\bibitem{goodfellow2013challenges}
I.~J. Goodfellow, D.~Erhan, P.~L. Carrier, A.~Courville, M.~Mirza, B.~Hamner,
  W.~Cukierski, Y.~Tang, D.~Thaler, D.-H. Lee \emph{et~al.}, ``Challenges in
  representation learning: A report on three machine learning contests,'' in
  \emph{International conference on neural information processing}.\hskip 1em
  plus 0.5em minus 0.4em\relax Springer, 2013, pp. 117--124.

\bibitem{dhall2011static}
A.~Dhall, R.~Goecke, S.~Lucey, and T.~Gedeon, ``Static facial expression
  analysis in tough conditions: Data, evaluation protocol and benchmark,'' in
  \emph{2011 IEEE International Conference on Computer Vision Workshops (ICCV
  Workshops)}.\hskip 1em plus 0.5em minus 0.4em\relax IEEE, 2011, pp.
  2106--2112.

\bibitem{zhang2016joint}
K.~Zhang, Z.~Zhang, Z.~Li, and Y.~Qiao, ``Joint face detection and alignment
  using multitask cascaded convolutional networks,'' \emph{IEEE signal
  processing letters}, vol.~23, no.~10, pp. 1499--1503, 2016.

\bibitem{jiang2022disentangling}
J.~Jiang and W.~Deng, ``Disentangling identity and pose for facial expression
  recognition,'' \emph{IEEE Transactions on Affective Computing}, 2022.

\bibitem{li2021adaptively}
H.~Li, N.~Wang, X.~Ding, X.~Yang, and X.~Gao, ``Adaptively learning facial
  expression representation via cf labels and distillation,'' \emph{IEEE
  Transactions on Image Processing}, vol.~30, pp. 2016--2028, 2021.

\bibitem{zhang2021learning}
W.~Zhang, X.~Ji, K.~Chen, Y.~Ding, and C.~Fan, ``Learning a facial expression
  embedding disentangled from identity,'' in \emph{CVPR}, 2021, pp. 6759--6768.

\bibitem{li2022crs}
H.~Li, N.~Wang, X.~Yang, and X.~Gao, ``Crs-cont: A well-trained general encoder
  for facial expression analysis,'' \emph{IEEE Transactions on Image
  Processing}, vol.~31, pp. 4637--4650, 2022.

\bibitem{acharya2018covariance}
D.~Acharya, Z.~Huang, D.~Pani~Paudel, and L.~Van~Gool, ``Covariance pooling for
  facial expression recognition,'' in \emph{Proceedings of the IEEE Conference
  on Computer Vision and Pattern Recognition Workshops}, 2018, pp. 367--374.

\bibitem{cai2018island}
J.~Cai, Z.~Meng, A.~S. Khan, Z.~Li, J.~O'Reilly, and Y.~Tong, ``Island loss for
  learning discriminative features in facial expression recognition,'' in
  \emph{2018 13th IEEE International Conference on Automatic Face \& Gesture
  Recognition (FG 2018)}.\hskip 1em plus 0.5em minus 0.4em\relax IEEE, 2018,
  pp. 302--309.

\bibitem{zeng2018facial}
J.~Zeng, S.~Shan, and X.~Chen, ``Facial expression recognition with
  inconsistently annotated datasets,'' in \emph{Proceedings of the European
  conference on computer vision (ECCV)}, 2018, pp. 222--237.

\bibitem{van2008visualizing}
L.~Van~der Maaten and G.~Hinton, ``Visualizing data using t-sne.''
  \emph{Journal of machine learning research}, vol.~9, no.~11, 2008.

\end{thebibliography}
}

\end{document}